\documentclass{article} % For LaTeX2e
\usepackage{iclr2027_conference,times}

\usepackage{amsmath,amsfonts,bm}

\def\eqref#1{equation~\ref{#1}}
\def\1{\bm{1}}

\def\vtheta{{\bm{\theta}}}

\def\ve{{\bm{e}}}

\def\vr{{\bm{r}}}

\def\vt{{\bm{t}}}

\def\vx{{\bm{x}}}

\def\vz{{\bm{z}}}

\DeclareMathAlphabet{\mathsfit}{\encodingdefault}{\sfdefault}{m}{sl}
\SetMathAlphabet{\mathsfit}{bold}{\encodingdefault}{\sfdefault}{bx}{n}

\def\sD{{\mathbb{D}}}
\DeclareMathOperator*{\argmin}{arg\,min}

\usepackage{hyperref}
\usepackage{url}
\usepackage{graphicx}
\usepackage{booktabs}
\usepackage{multirow}
\usepackage{enumitem}
\usepackage{xspace}
\usepackage{xcolor}
\usepackage{textcomp} % needed for upquote=true in listings

\newcommand{\mysection}[1]{{\textbf{#1.}}\xspace}

\usepackage{iftex}
\ifPDFTeX
  \usepackage[T1]{fontenc}
  \newcommand{\appendixfont}{\ttfamily}
\else
  \usepackage[no-math]{fontspec}
  \newfontfamily\appendixfont{DejaVuSansMono.ttf}[
    BoldFont=DejaVuSansMono-Bold.ttf,Scale=0.85]
\fi

\usepackage{listings}
\usepackage[skins,breakable,listings]{tcolorbox}

\definecolor{AppendixBlue}{HTML}{334E68}
\definecolor{AppendixRule}{HTML}{CDD5DE}
\definecolor{AppendixBackground}{HTML}{FAFBFC}
\definecolor{AppendixComment}{HTML}{47594D}

\lstdefinestyle{appendixcode}{
  language=Python,
  basicstyle=\appendixfont\fontsize{8}{10}\selectfont,
  keywordstyle=\color{AppendixBlue}\bfseries,
  commentstyle=\color{AppendixComment},
  stringstyle=\color{black!75},
  numbers=none,
  numberstyle=\tiny\color{black!40},
  numbersep=8pt,
  xleftmargin=0pt,
  columns=fullflexible,
  keepspaces=true,
  showstringspaces=false,
  breaklines=true,
  breakatwhitespace=false,
  breakindent=12pt,
  tabsize=4,
  literate={—}{{---}}3 {–}{{--}}2
           {×}{{$\times$}}1 {≈}{{$\approx$}}1
}

\lstdefinestyle{appendixprompt}{
  style=appendixcode,
  language={},
  numbers=none,
  xleftmargin=0pt,
  basicstyle=\appendixfont\fontsize{8}{10}\selectfont,
  breakatwhitespace=true,
  breakindent=0pt
}

\tcbset{appendixbox/.style={
  enhanced,
  breakable,
  listing only,
  colback=AppendixBackground,
  colframe=AppendixRule,
  colbacktitle=AppendixBlue!8,
  coltitle=AppendixBlue,
  fonttitle=\small\bfseries,
  boxrule=0.5pt,
  arc=1.5mm,
  left=8pt,right=8pt,top=6pt,bottom=6pt,
  toptitle=5pt,bottomtitle=5pt,
  before skip=10pt,after skip=12pt
}}

\newtcblisting[auto counter]{programbox}[2]{
  appendixbox,
  title={Program~\thetcbcounter. #1},
  title after break={Program~\thetcbcounter. #1 (continued)},
  label={#2},
  listing options={style=appendixcode}
}

\newtcblisting[auto counter]{promptbox}[2]{
  appendixbox,
  title={Prompt~\thetcbcounter. #1},
  title after break={Prompt~\thetcbcounter. #1 (continued)},
  label={#2},
  listing options={style=appendixprompt}
}

\title{Evolving Semantic Indexes for Generative Retrieval}

\author{
Jinming Nian$^{1}$\thanks{Work done during an internship at Coupang.}\quad
Jyun-Yu Jiang$^{2}$\quad
Henry Tsai$^{2}$\quad
Yi Fang$^{1}$\quad
Zhirong Liang$^{2}$ \\[4pt]
$^{1}$Santa Clara University \qquad $^{2}$Coupang \\[2pt]
\small\texttt{jnian@scu.edu},\;
\texttt{jyunyu.jiang@gmail.com}, \\
\small\texttt{henry\_tsai@haas.executivealumni.berkeley.edu}, \\
\small\texttt{yfang@scu.edu},\;
\texttt{zhliang@coupang.com}
}

\iclrfinalcopy % Shows author names and removes line numbers.
\begin{document}

\maketitle
\lhead{Preprint. Under review.} % Overrides the "Published as a conference paper at ICLR 2027" header set by \maketitle.

\begin{abstract}
    Generative retrieval (GR) casts search and recommendation as autoregressive generation of discrete semantic identifiers (SIDs). SID design largely determines the quality of GR results: SID models, collision handling mechanisms, decoding frameworks, and training recipes can each change retrieval performance significantly. However, these discrete and interdependent design choices cannot be optimized end-to-end using gradients from downstream retrieval and therefore remain largely designed by hand. In this paper, we introduce a framework for automatically improving these components through LLM-driven evolutionary search on the SID framework, which to the best of our knowledge is the first application of LLM-driven evolutionary program search to GR. A locally servable, open-weight 31B language model is used to mutate executable GR components, so no proprietary LLM API is used at any point, and a 1M-parameter proxy decoder provides fast downstream feedback in place of repeatedly training the 1B target generator, which would otherwise make evolutionary search computationally infeasible. Across 21 SID programs on NQ320K, proxy and target Recall@100 exhibit strong rank correlation ($\rho=0.90$). Guided by this proxy, evolution consistently improves the seed GR on NQ320K, MSMARCO300K, and a multilingual e-commerce dataset, under a training recipe held fixed across conditions. On NQ320K, jointly evolving SID and collision handling improves Recall@1 from 55.8\% to 59.3\% and Recall@100 from 85.8\% to 90.6\%. Our component study reveals that proxy reliability depends on which component is evolved. Improvements transfer reliably when evolving the semantic index, but transfer less reliably when evolving the generator's training objective or the full GR pipeline. These results suggest that LLM-driven evolutionary search is a practical method for improving GR design and show that carefully assessing proxy reliability is essential for successful GR pipeline optimization.
\end{abstract}

\section{Introduction}
\label{sec:introduction}

Generative retrieval (GR) casts document retrieval as an autoregressive generation task. Given a query, the model is trained to generate the document identifier which maps back to a document~\citep{DSI, NCI}. In GR systems that are based on semantic identifiers (SIDs), documents are represented by short sequences of discrete tokens, often constructed from their semantic embeddings~\citep{TIGER, grid}. These identifiers define the prediction problem for the generator. During autoregressive decoding, documents that share an SID prefix share early generation decisions, which means a relevant document will become unreachable if an SID token in its prefix is discarded during beam search. SID design therefore determines both what the GR model must learn and how it can reach relevant documents during decoding~\citep{10.1145/3589334.3645477}. Improving a generative retriever thus requires careful design of SID generation. This design could involve several interdependent procedures: (1) an index builder assigns SIDs to documents; (2) collision handling distinguishes documents that were assigned the same SID; (3) a training objective teaches the generator to predict these identifiers; (4) generative decoding searches over valid SID sequences. Each procedure affects how the others operate. For example, changing the SID structure reforms the prediction targets, while resolving collisions may induce longer SID sequences for the generator to learn. Existing methods have shown that SID and generative retrieval models can be learned jointly within predefined architectures~\citep{GenRet, ASI}. However, ordinary backpropagation does not directly choose among different implementations of SID construction, collision handling, training, and decoding. Optimizing these executable procedures presents a broader, discrete search problem. To the best of our knowledge, no prior work has applied LLM-driven evolutionary program search to generative retrieval, and this paper is the first to show that doing so improves a GR pipeline.

\begin{figure}[t]
  \centering
  \includegraphics[width=1.0\linewidth]{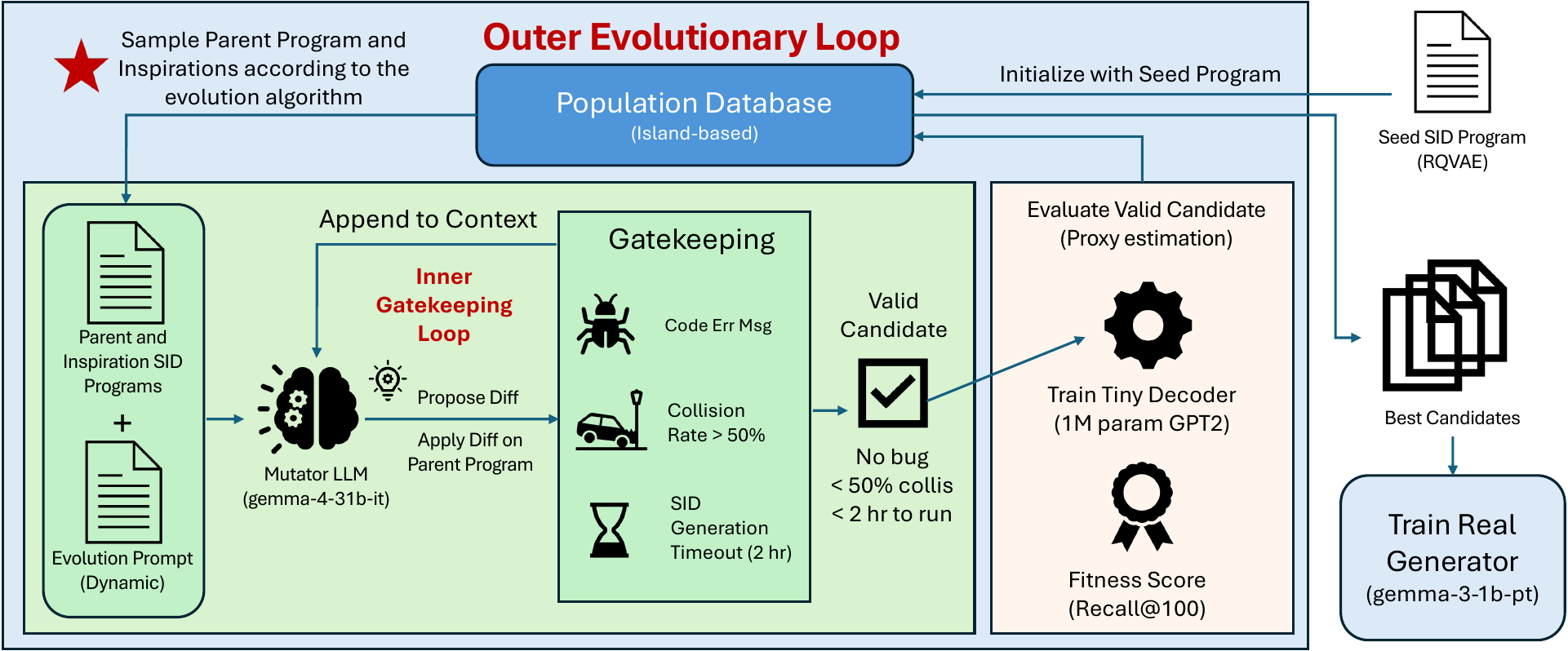}
  \caption{Our two-tiered evolutionary framework. The outer loop manages the population database, selects parent programs, and constructs feedback for the mutator LLM. The inner loop gatekeeps and refines candidate programs, approximating fitness with a tiny decoder.}
  \label{fig:method}
\end{figure}

LLM-driven evolutionary search offers a way to explore this search space by proposing code mutations and retaining programs that perform well on an execution-based evaluation~\citep{funsearch,novikov2025alphaevolvecodingagentscientific}. While recent work has applied evolutionary program search to discover lexical retrieval~\citep{nian2026rankevolve} and reranking pipelines~\citep{lin2026component}, directly applying this approach to generative retrieval faces a substantial evaluation bottleneck. Evaluating each candidate index requires training the downstream generator and measuring its retrieval performance. In our setting, a target generator training run takes five hours on an H200 GPU, whereas training a proxy takes 20 minutes. Using the proxy therefore reduces the per-candidate evaluation time by 15$\times$, making it practical to evaluate many more candidate programs within a fixed computational budget. The central challenge is therefore to obtain inexpensive feedback that remains informative for guiding evolutionary search. In addition, recent discovery systems often use frontier models as the mutator~\citep{qu2026coral,virk2026swarmresearch,adaevolve}, which is expensive and less transparent. We run the entire search with an open-weight model that fits on a single GPU. We do not claim that a 31B mutator LLM is sufficient for program search in general, only that it was sufficient for evolving GR to outperform the seed.

As shown in Figure~\ref{fig:method}, we introduce a framework for evolving executable generative retrieval components using a small retrieval proxy and a locally served, open-weight mutator. Specifically, we use \texttt{gemma-4-31b-it}~\citep{gemmateam2026gemma4}, which can be served locally on a single H200 via vLLM~\citep{vllm}. Building on the AdaEvolve search algorithm~\citep{adaevolve}, the mutator proposes \texttt{SEARCH/REPLACE} edits to the program under fixed input and output constraints. For SID generation, the program receives document and query embeddings together with training relevance labels and returns an SID for each document. Evolution can change the procedure implementing this mapping. A tiny autoregressive decoder with approximately 1M non-embedding parameters is trained and tested on the training dataset only, to provide proxy retrieval performance feedback for candidate program evaluation. Programs are selected and evolved using proxy feedback alone, and in the end the best candidate is fully evaluated by training the \texttt{google/gemma-3-1b-pt}~\citep{gemmateam2025gemma3} target generator on the training dataset and measuring the retrieval performance on the reserved test set from its final checkpoint.
Across NQ320K~\citep{NQ}, MSMARCO300K~\citep{MSMARCO}, and a multilingual e-commerce dataset, the selected evolved SID programs improve their corresponding seed programs when the SIDs they produce are used to train the target 1B generator. On NQ320K, jointly evolving SID construction and collision handling improves Recall@1 from 55.8\% to 59.3\% and Recall@100 from 85.8\% to 90.6\%. Our component study also identifies a boundary of proxy-guided search. Improvements transfer more consistently for the index-construction programs, whereas changes to the generator's training objective or the full GR pipeline can improve proxy performance while reducing target performance. This distinction suggests that the usefulness of a small proxy evaluator depends on the artifacts being optimized. A shared SID structure can benefit both proxy and target generators, while a training objective that the proxy benefits from may not be suitable for the much larger target decoder. We state the scope of our claims plainly. We do not claim state-of-the-art GR, nor that GR dominates lexical or dense retrieval. Our contributions are as follows:
\begin{itemize}[leftmargin=*]
    \item We introduce a framework for evolving executable generative retrieval procedures with a tiny proxy decoder and a locally served 31B language model mutator, the first such framework for generative retrieval and one that uses no proprietary LLM API.
    \item We demonstrate retrieval improvements over matched seed systems across three retrieval datasets, observing substantial gains from jointly evolving SID construction and collision handling.
    \item We show that proxy-to-target transfer depends on the component being evolved, with semantic index improvements transferring more consistently than decoder training objective changes.
\end{itemize}

\section{Generative Retrieval}
\label{sec:generative_retrieval}

Generative retrieval maps a query to a document by generating its identifier~\citep{DSI,NCI}. Let $\sD$ denote the document corpus. Each document $d \in \sD$ is assigned a semantic identifier $\vz(d) = (z_1(d),\ldots,z_{L(d)}(d))$, where $L(d)$ is the number of tokens in its identifier. Given query text $q$, an autoregressive generator with parameters $\vtheta$ models
\begin{equation}
    P_{\vtheta}(\vz(d)\mid q)
    =
    \prod_{\ell=1}^{L(d)}
    P_{\vtheta}
    \bigl(z_\ell(d)\mid q,\vz_{<\ell}(d)\bigr),
    \label{eq:sid_likelihood}
\end{equation}
where $\vz_{<\ell}(d)$ denotes the preceding identifier tokens. We train the generator on relevant query--SID pairs using teacher-forced next-token cross-entropy loss.

Inference traces a prefix trie of the valid SID sequences. Constrained beam search only permits child trie nodes as next tokens. In the end, the completed SIDs are mapped back to documents. Documents sharing the same SID prefix share early decoding decisions, so pruning a prefix early removes every document beneath it from the remaining search. The SID index therefore determines both the prediction targets during training and the branching structure during retrieval.

\subsection{Semantic Indexing}
\label{sec:semantic_indexing}

Our seed program combines RQVAE-based SID construction with TIGER-style collision handling~\citep{RQVAE,TIGER}. We first encode document text using a fixed Contriever or mContriever encoder~\citep{izacard2022unsupervised}. These embeddings provide the input to the index builder. The generative retriever only receives query text.

\mysection{SID construction} The seed constructs SIDs using residual quantization. Let $\vx(d)$ be the fixed embedding of document $d$ and $\vr_0(d)$ its latent representation produced by the RQVAE encoder. At each level $\ell$, the quantizer selects a codebook vector and subtracts it from the current residual:
\begin{equation}
    z_\ell(d)
    = \argmin_k
       \left\|\vr_{\ell-1}(d)-\ve_{\ell,k}\right\|_2^2,
    \qquad
    \vr_\ell(d)
    = \vr_{\ell-1}(d)-\ve_{\ell,z_\ell(d)},
\label{eq:residual_quantization}
\end{equation}
where $\ve_{\ell,k}$ is the $k$-th vector in the codebook at level $\ell$. After $M$ levels, the selected indices $(z_1(d),\ldots,z_M(d))$ form the semantic part of the identifier. Following \citet{RQVAE}, the RQVAE is trained with reconstruction and quantization losses to preserve information from the document embeddings. These losses do not directly optimize downstream retrieval performance.

\mysection{Collision handling}
SID construction commonly assigns the same SID sequence to multiple documents. Following TIGER~\citep{TIGER}, the seed program adds a unique disambiguation token $z_{M+1}(d)$ within each group of documents that were assigned the same SID sequence:
\begin{equation}
    \vz(d)
    = \bigl(z_1(d),\ldots,z_M(d),z_{M+1}(d)\bigr),
    \qquad L(d)=M+1.
\label{eq:complete_sid}
\end{equation}

SID construction and collision handling define the generator prediction targets. Semantic tokens determine which documents share a prefix, while the final token distinguishes documents with identical semantic codes. Because changing the quantizer also changes the collision groups, we study both evolving SID construction with fixed collision handling and evolving the two components jointly.

\section{Proxy-Guided Evolutionary Search}
\label{sec:proxy_method}

We mainly evolve the program that constructs SIDs and use a small generative retriever to evaluate the SIDs it produces. Each candidate program is evaluated by executing the program to obtain document-to-SID mappings, training the small generator on the relevant query--SID pairs, and measuring its retrieval performance. This feedback serves as a proxy that guides evolutionary program search without training the full 1B target generator at every step.

\mysection{Executable search space} When evolving semantic indexes, the candidate program receives fixed corpus document embeddings, training query embeddings, and training relevance pairs. The program returns an integer array containing one SID per document, aligned to the input documents. The input and output interface is fixed, while the procedure implementing this mapping is mutable. For example, evolution can change the quantization process or codebook initialization, or even switch to an entirely different method. The input embeddings and relevance labels are fixed and come from the training data. The mutator LLM only has access to the SID generation code and performance feedback. The mutation prompts (Appendix~\ref{app:mutator_prompts}) encourage creativity but also include suggested algorithm families, such as alternative quantization and clustering methods. Test data is entirely unreachable to the mutator LLM.

For SID-only search, collision handling is held fixed. For joint SID and collision handling search, the mutable program determines how documents with identical SIDs are distinguished. We also study broader search spaces for changes to the training objective of generative retrievers, generator decoding logic, and collision-handling-only changes. Full-pipeline search is also studied, where SID construction, collision handling, generator training, and decoding can all be changed. These scopes allow us to examine how the choice of mutable component affects proxy-guided search.

\mysection{Proxy generative retriever} While SID collision rate and per-codebook perplexity can describe desired properties of SIDs, they do not directly measure the SIDs' usefulness for retrieval. Across 21 evaluated NQ320K SID programs, the collision rate logged during evolution is negatively associated with final target Recall@100 (Spearman $\rho=-0.855$). Although this makes collision rate a usable signal, it cannot serve as the fitness function because it is too easy to hack. For example, random assignments can produce few collisions and balanced code usage without providing meaningful semantic structure for the downstream generator. We therefore evaluate candidate programs through downstream retrieval, using a small generator to assess how well their identifiers can be learned and retrieved. Our proxy is a two-layer GPT-2~\citep{radford2019language} model with approximately 1M non-embedding parameters (7M including token embeddings). Like the target 1B generator, it receives query text and predicts relevant documents' SIDs using constrained decoding.

To keep search blind to test data, we randomly split each dataset's original training relevance pairs into $\mathcal{R}_{\mathrm{proxy}}$ (two thirds) and $\mathcal{R}_{\mathrm{feedback}}$ (one third). The split is over query--document pairs, so a query with several relevant documents can contribute pairs to both sides. During search, both the candidate program and the proxy retriever $g_{\mathrm{P}}(c)$ see only $\mathcal{R}_{\mathrm{proxy}}$, and the candidate's fitness is
\begin{equation}
    F_{\mathrm{P}}(c)
    = \operatorname{Recall@100}
      \bigl(g_{\mathrm{P}}(c);\mathcal{R}_{\mathrm{feedback}}\bigr).
    \label{eq:proxy_fitness}
\end{equation}
The proxy measures how well a small generator can learn and retrieve the candidate's SIDs. Its scores do not need to match the target generator, but they must provide useful preferences among programs in the search space. Whether these preferences transfer to the target generator is an empirical question, which we explore in Section~\ref{sec:proxy_transfer}.

\mysection{Program mutation and execution}
As shown in Figure~\ref{fig:method}, we use AdaEvolve~\citep{adaevolve} to maintain an island-based population of programs and control exploration and exploitation. At each outer step, the search controller selects a parent program and additional programs to provide context for mutation. We use \texttt{gemma-4-31b-it}~\citep{gemmateam2026gemma4}, served locally with vLLM~\citep{vllm} on a single H200, to propose \texttt{SEARCH/REPLACE} edits within the marked mutable regions of the parent program. The proposed edits are applied to the parent, and the resulting candidate program is executed. We check that the candidate executes successfully and passes the collision gate before training the proxy. For SID-only search, we reject candidates with a collision rate above 50\%, and if a candidate is accepted, we apply TIGER-style collision handling. Candidate execution is limited to two hours. When a proposal fails, the mutator receives the execution error, collision diagnostics, or timeout message and is asked to revise its edits to the parent. We allow up to 20 retries before abandoning the proposal and moving to the next outer step.

\section{Experimental Setup}
\label{sec:experiments}

Our experiments examine whether programs selected with a small, cheap proxy improve the final 1B generator, and how this transfer depends on the component being evolved. We compare evolved programs against the seed under a fixed target training recipe, measure proxy--target agreement, and inspect the evolved programs and computational cost.

\mysection{Datasets and Baselines}
We evaluate on NQ320K~\citep{NQ} and MSMARCO300K~\citep{MSMARCO}, which are both reduced collections of the original datasets, following prior work~\citep{DSI,10.1145/3603167,zhou-etal-2023-enhancing-generative,10.1145/3726302.3730023}. We also evaluate on a proprietary multilingual dataset from a large e-commerce company. The NQ320K collection contains 109,739 documents and 7,830 evaluation queries. The MSMARCO300K subset contains 319,927 documents and 808 evaluation queries. The e-commerce corpus contains millions of distinct purchased items collected over one month. We use two random seeds to downsample the purchase queries into two equal-sized training sets. Both are tested on 25,000 purchase queries from the day following that month, stratified-sampled by purchase frequency. Items are represented by title, description, and brand, with purchases as query--item relevance labels.

We use fixed Contriever~\citep{izacard2022unsupervised} embeddings for NQ and MSMARCO and mContriever embeddings for e-commerce because most of its text is not in English. Keeping the encoder fixed makes the seed and evolved SID programs operate on the same input representations, which isolates the effect of SID structure. In Table~\ref{tab:main_retrieval}, the only difference between the Seed GR and SID evolution rows is the SID program (SID construction and, where evolved, collision handling). The SID construction program receives document and query embeddings together with training relevance pairs, and the generative retriever receives query text, the resulting SIDs, and relevance pairs. Search uses only the original training relevance pairs, split as in Section~\ref{sec:proxy_method}, with no synthetic or test queries. For final training, the best program by proxy fitness is rerun on all original training pairs to produce SIDs, and the target generator is trained on the same pairs. For NQ and MSMARCO only, both steps also include one synthetic query per corpus document generated by \texttt{gemma-4-31b-it}~\citep{gemmateam2026gemma4}, following prior work~\citep{zhou2022ultron,zhou-etal-2023-enhancing-generative,zhang2024tsgen,10.1145/3726302.3730023}, which typically generates ten per document with a weaker generator. E-commerce uses no synthetic queries because purchase data is plentiful. All reported numbers are on the held-out test queries, which neither stage ever sees. Our primary comparison is against the generative retrieval pipeline using the seed program (Seed GR), which uses RQVAE and the fixed collision handling described in Section~\ref{sec:semantic_indexing}. We also report BM25 and Contriever performance as references rather than competitors.

\mysection{Program Search and Evaluation Configuration}
The seed program uses three residual codebooks with 256 entries each and a 256-dimensional latent representation. We train the seed RQVAE~\citep{RQVAE,TIGER} for 100 epochs using AdamW with learning rate $10^{-4}$, batch size 2,048, weight decay 0.01, and commitment coefficient 0.01. The proxy is a two-layer GPT-2 decoder with approximately 1M non-embedding parameters, or 7M including token embeddings. AdaEvolve~\citep{adaevolve} maintains the program population with 3 islands, and \texttt{gemma-4-31b-it} proposes mutations through vLLM~\citep{vllm} on one H200. Candidate execution is limited to two hours, with up to 20 inner-loop retries. SID-only candidates must pass a 50\% collision-rate gate before fixed collision handling is applied. Joint SID and collision handling candidates must pass a 5\% gate on their returned SIDs. After each evolution step, the proxy GPT-2 is trained with cross-entropy loss on SID tokens and EOS for 10 epochs using AdamW with learning rate $5\times 10^{-4}$, batch size 512, and weight decay 0.01.

We study several evolution regimes: \emph{SID-only} evolves SID construction while retaining TIGER collision handling, whereas \emph{joint SID-collision} updates both simultaneously. On NQ, we additionally evaluate \emph{collision evolution} (fixing the seed SID program), \emph{sequential SID-collision} (evolving collision handling after fixing an evolved SID constructor), \emph{objective evolution} (modifying the downstream generator's loss, distinct from the SID constructor's loss), and \emph{full-pipeline evolution} (spanning SID construction, collision handling, generator training, and decoding).

We evaluate the best program from each evolution run with \texttt{gemma-3-1b-pt}~\citep{gemmateam2025gemma3}, which we refer to as the ``target'' generator. We use the pretrained rather than the instruction-tuned checkpoint because it carries general pretraining knowledge without having been adapted to any downstream task, which makes it a clean base for learning the GR objective. It is always trained with AdamW for 3 epochs with learning rate $5\times 10^{-5}$, batch size 32, weight decay 0.01, and warmup fraction 0.05. On an H200 GPU, a proxy run takes around 20 minutes, compared with five hours for the target generator. For 100 candidate evaluations, these runtimes correspond to 33 hours of proxy computation versus 500 hours of target evaluation. LLM mutation, SID construction, and failed candidate executions incur additional costs. We report the target generator's Recall@1, Recall@5, Recall@10, Recall@100, and MRR@10, all as percentages.

\begin{table}[t]
\centering
\caption{\textbf{Overall results.} ``SID evolution'' reports the target's performance when trained on SIDs generated by the best SID program found through evolutionary search, where the best program is selected using the proxy's Recall@100. Values with $\pm$ are means and standard deviations across three independent training runs; for e-commerce, they pool three runs on each of the two training sets (six runs). For the proprietary e-commerce dataset, absolute scores are withheld; we report absolute gains over BM25 in percentage points. Bold indicates the highest mean for each metric within a dataset.}
\label{tab:main_retrieval}

\resizebox{\textwidth}{!}{%
\begin{tabular}{llccccc}
\toprule
Dataset & Method & R@1 (\%) & R@5 (\%) & R@10 (\%) & R@100 (\%) & MRR@10 (\%) \\
\midrule
\multirow{4}{*}{NQ320K}
 & BM25 & 32.08 & 54.51 & 62.35 & 83.69 & 41.62 \\
 & Contriever & 36.18 & 62.45 & 71.37 & 89.41 & 47.36 \\
 & Seed GR & $55.82\pm0.04$ & $71.12\pm0.53$ & $75.27\pm0.44$ & $85.82\pm0.22$ & $62.50\pm0.19$ \\
 & SID evolution & $\mathbf{59.28\pm0.06}$ & $\mathbf{76.44\pm0.18}$ & $\mathbf{81.05\pm0.09}$ & $\mathbf{90.57\pm0.20}$ & $\mathbf{66.74\pm0.08}$ \\
\midrule
\multirow{4}{*}{MSMARCO300K}
 & BM25 & 18.94 & \textbf{47.52} & \textbf{60.64} & \textbf{86.63} & \textbf{31.32} \\
 & Contriever & 12.38 & 30.82 & 41.58 & 73.64 & 20.56 \\
 & Seed GR & $19.43\pm0.18$ & $40.97\pm0.35$ & $51.49\pm0.88$ & $79.70\pm0.18$ & $28.61\pm0.11$ \\
 & SID evolution & $\mathbf{20.48\pm0.44}$ & $41.96\pm1.05$ & $55.01\pm0.26$ & $82.55\pm1.40$ & $30.09\pm0.06$ \\
\midrule
\multirow{4}{*}{\shortstack[l]{E-commerce\\(gain vs.\ BM25)}}
 & BM25 & ref. & ref. & ref. & ref. & ref. \\
 & mContriever & $+6.61$ & $+14.06$ & $+17.39$ & $+29.13$ & $+9.73$ \\
 & Seed GR & $+8.06\pm0.60$ & $+20.36\pm1.00$ & $+27.18\pm0.84$ & $+49.01\pm0.45$ & $+13.37\pm0.73$ \\
 & SID evolution & $\mathbf{+8.56\pm0.68}$ & $\mathbf{+21.40\pm0.85}$ & $\mathbf{+28.47\pm0.74}$ & $\mathbf{+50.66\pm0.47}$ & $\mathbf{+14.09\pm0.75}$ \\
\bottomrule
\end{tabular}}
\end{table}

\section{Results}

\subsection{Evolved Semantic Indexes}
Table~\ref{tab:main_retrieval} shows improvements over Seed GR on all three datasets. SID evolution increases Recall@100 from 85.82\% to 90.57\% on NQ and from 79.70\% to 82.55\% on MSMARCO, and raises the e-commerce Recall@100 gain over BM25 from $+49.01$ to $+50.66$ points. Recall@1 and MRR@10 also improve on all three datasets, indicating that although evolutionary search optimizes Recall@100, its benefits extend to early ranking quality rather than only retrieving additional relevant documents near the end of the list; on e-commerce, however, these early-rank gains are within one standard deviation. For NQ and MSMARCO, the best SID evolution program comes from jointly evolving SID construction and collision handling. For e-commerce, it comes from evolving SID construction only. On NQ, the seed underperforms Contriever on Recall@100, yet a different SID structure is sufficient for the final generator to outperform Contriever, while everything else, from data to the target generator and training recipe, remains the same. On MSMARCO, evolved GR exceeds BM25 at rank one, but BM25 remains stronger at larger recall cutoffs and in MRR@10. We do not claim that the evolved GR pipeline achieves state-of-the-art results, nor that GR dominates lexical or dense retrieval. Instead, we argue that evolutionary search on the semantic index can reliably improve a well-established GR pipeline whose every other component is held fixed. While improving the semantic index improves retrieval performance, it does not eliminate the limitations of the general GR pipeline. On the e-commerce dataset, where relevance is defined by purchase rather than by whether the document answers the query as in NQ and MSMARCO, GR outperforms BM25 by a large margin. This is expected because the purchased product does not necessarily contain the query terms in its title, brand, or description. Search queries are also often short and ambiguous, which makes it hard for lexical retrieval to compete with dense and generative retrieval.

\subsection{Proxy Transfer Depends on What is Evolved}
\label{sec:proxy_transfer}

\begin{figure}[t]
\centering
\begin{minipage}[t]{0.49\linewidth}
\centering
\includegraphics[width=\linewidth]{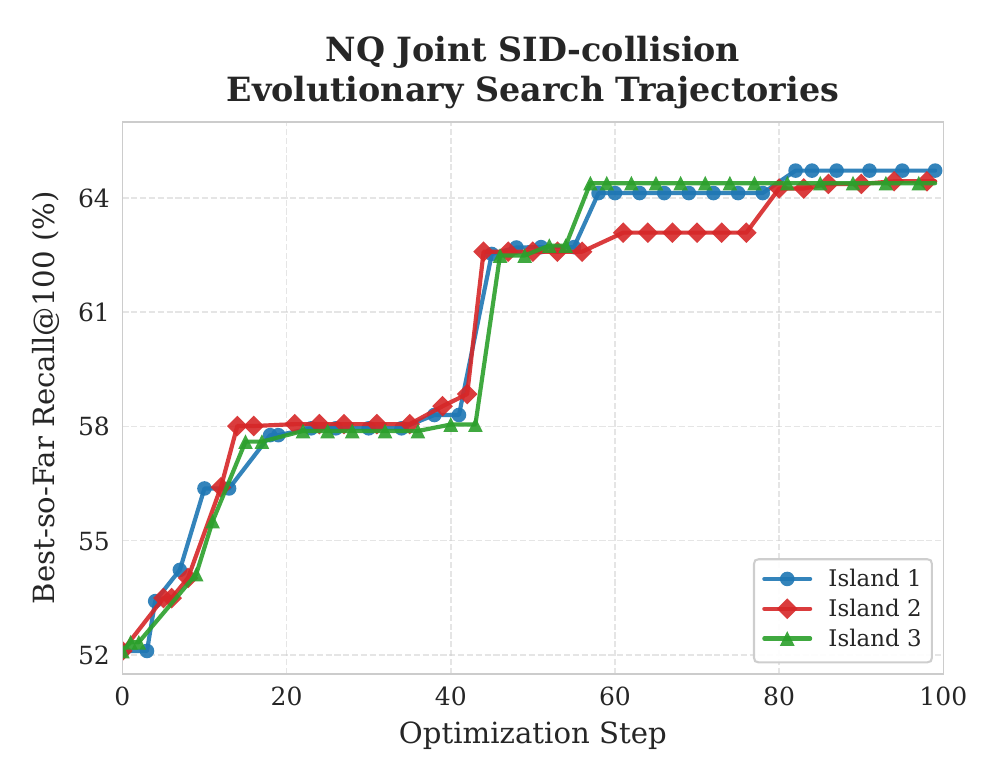}
\par\smallskip
{\footnotesize (a) Three-island search progression}
\end{minipage}\hfill
\begin{minipage}[t]{0.49\linewidth}
\centering
\includegraphics[width=\linewidth]{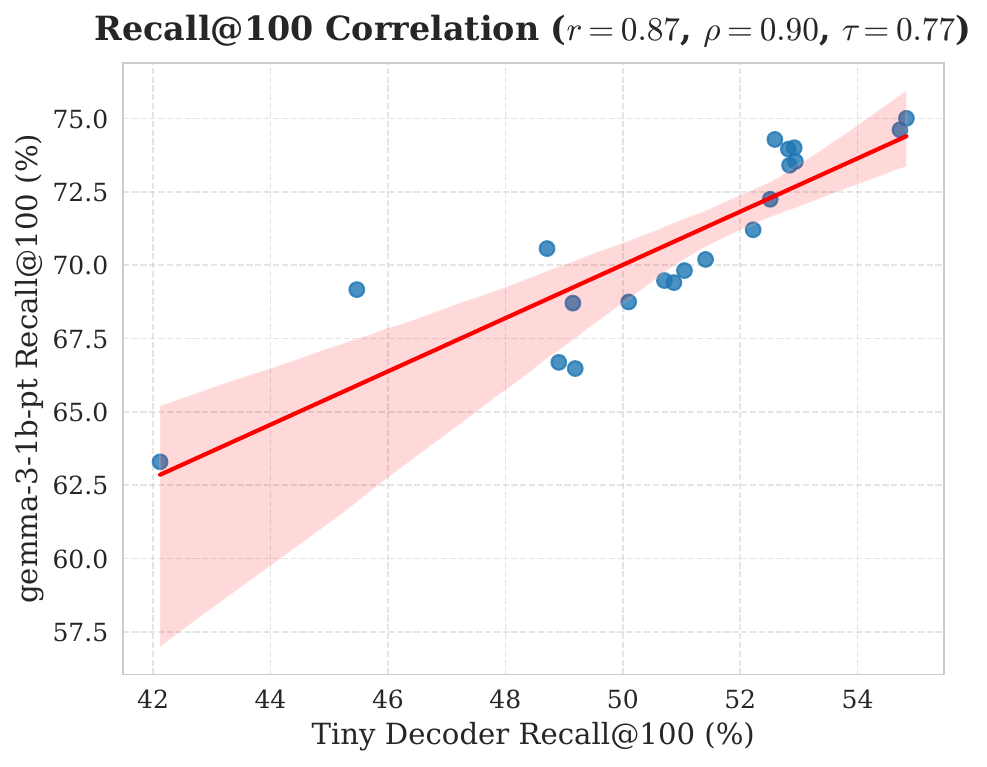}
\par\smallskip
{\footnotesize (b) Proxy--target correlation}
\end{minipage}
\caption{(a) Joint SID-collision search on NQ with three islands; island-based evolution is a main feature of AdaEvolve. Each curve connects the recorded best-so-far proxy Recall@100 values for one island. The $x$-axis is the global evolution step. (b) A separate correlation experiment over 21 joint SID-collision programs pairs proxy and target Recall@100, with Spearman $\rho=0.90$. Both models are trained without synthetic queries, so absolute target scores are lower than in Table~\ref{tab:main_retrieval}. The red shaded region represents the 95\% confidence interval.}
\label{fig:proxy_index_transfer}
\label{fig:island_progression}
\end{figure}

Figure~\ref{fig:island_progression}(a) shows the optimization trajectory of the run that produced the best joint SID-collision program for NQ. The LLM-driven evolutionary search steadily improves the defined fitness score. If the evaluator were exactly faithful to the real task, the program it selects would be the best program for that task. In practice, an exact evaluator is often too expensive. In our case, the true evaluation is training the target \texttt{gemma-3-1b-pt} on the SIDs produced by each mutated program, which takes five hours; a 100-step search would cost nearly 500 hours, or about 21 days. In production e-commerce settings, where the generator is trained on a year of purchase data, a single training run can take weeks. This motivates a lightweight proxy evaluator, whose usefulness depends on how well its rankings agree with the downstream task. Figure~\ref{fig:proxy_index_transfer}(b) examines 21 programs from a separate joint SID-collision calibration experiment: 20 candidates sampled at random from across the optimization, plus the seed. We train both the 1M-parameter proxy and the 1B-parameter target generator on the SIDs produced by these 21 programs on NQ, without synthetic queries. Proxy and target Recall@100 have a strong positive rank correlation (Spearman $\rho=0.90$). The proxy need not reproduce the target's absolute score to be useful. As long as it favors semantic index changes that also help the target generator, evolutionary search can be effective.

\begin{table}[t]
\centering
\caption{\textbf{Transfer across search scopes.} Selected NQ programs, grouped by whether the generator objective can change. Scores are percentages; $\Delta$ is target R@100 relative to Seed GR, in percentage points. Proxy values are single search evaluations; target values summarize three runs, except full-pipeline evolution (two). Objective and full-pipeline searches start at a proxy R@100 of 52.10\%. Searches have different initializations and completed budgets, so this compares observed outcomes rather than isolated component effects.}
\label{tab:scope_nq}
\begin{tabular}{lrrrr}
\toprule
Search scope & Proxy R@100 (\%) & Target R@1 (\%) & Target R@100 (\%) & $\Delta$ \\
\midrule
Seed GR & 52.31 & $55.82\pm0.04$ & $85.82\pm0.22$ & --- \\
\midrule
\multicolumn{5}{l}{\textit{Evolving the generator training objective}} \\
Objective evolution & 53.80 & $53.51\pm0.38$ & $84.46\pm0.33$ & $-1.36$ \\
Full-pipeline evolution & 62.85 & $53.85\pm0.50$ & $76.12\pm0.21$ & $-9.70$ \\
\midrule
\multicolumn{5}{l}{\textit{Evolving the semantic index (generator objective unchanged)}} \\
SID-only evolution & 61.47 & $56.33\pm0.40$ & $87.40\pm0.20$ & $+1.58$ \\
Collision evolution & 52.69 & $55.79\pm0.10$ & $85.72\pm0.37$ & $-0.10$ \\
Sequential SID--collision & 61.30 & $55.96\pm0.10$ & $87.25\pm0.21$ & $+1.43$ \\
Joint SID--collision & \textbf{64.39} & $\mathbf{59.28\pm0.06}$ & $\mathbf{90.57\pm0.20}$ & $\mathbf{+4.75}$ \\
\bottomrule
\end{tabular}
\vspace{-18pt}
\end{table}

The contrast in Table~\ref{tab:scope_nq} shows why this evidence cannot be extended to every mutable component. Once the generator's training objective becomes mutable, the proxy becomes not only less informative but misleading. Objective evolution increases proxy Recall@100 from its initialization of 52.10\% to 53.80\%, but target Recall@100 falls to 84.46\%. Full-pipeline evolution also mutates the generator's objective. It reaches a much higher proxy score of 62.85\% while reducing target Recall@100 to 76.12\%. This transfer failure extends beyond the selected examples in Table~\ref{tab:scope_nq}. Across five distinct evolved programs (three objective-only and two full-pipeline programs), all five improve proxy fitness over initialization while regressing on target recall relative to Seed GR, and proxy and target Recall@100 are inversely ranked (Spearman $\rho=-0.80$, $n=5$; too few programs for a significance claim). The main distinction is \emph{what} the generators learn versus \emph{how} they learn it. Semantic index evolution changes the SID structures shared by both generators. Objective evolution changes the optimization signal, whose effect can depend on model architecture, pretraining, and training recipe. An objective highly optimized for the 1M-parameter proxy does not necessarily transfer to the 1B-parameter target. Full-pipeline evolution, which can change both the objective and the semantic index, did not produce positive results either. Isolating per-component credit assignment during multi-component evolution is left to future work~\citep{lin2026component}. The finding we are sharing here is that proxy transfer depends on the allowed search region. 

As shown in Table~\ref{tab:scope_nq}, SID-only evolution improves target retrieval, while collision-only evolution leaves it close to the seed at 85.72\% versus 85.82\% Recall@100. Joint evolution provides the largest gain. Evolving collision handling after fixing an evolved SID constructor reaches 87.25\%, with no additional mean improvement over SID-only evolution at 87.40\%. Joint evolution reaches 90.57\%, exceeding the sequential result by 3.32 points. This pattern is consistent with a dependence between the two procedures. SID construction determines which documents share an identifier and thus the groups within which collision handling must distinguish documents. Holding that partition fixed restricts the collision procedure's choices, while joint search can change both the groups and the suffix targets. The NQ results suggest value in this coordinated design, although different search budgets, gates, and selected programs prevent attributing the gain to coordination alone.

\section{The Evolved SID Program}
\label{sec:nq_program_analysis}

We examine the joint SID and collision handling program evolved on NQ, which reaches a proxy Recall@100 of 64.39\% and raises target Recall@100 from 85.82\% to 90.57\% (Program~\ref{lst:nq_joint_program}). The program retains the RQVAE structure and the four-token SID of Equation~\ref{eq:complete_sid}, with $M=3$ semantic tokens and one collision suffix. Its changes mainly concern how the semantic codes are assigned and trained, and how the suffix is constructed.

\mysection{Direction-based residual quantization} The evolved program selects the code with maximum cosine similarity to the current residual, replacing the squared-Euclidean assignment in Equation~\ref{eq:residual_quantization}. It still subtracts the selected, unnormalized code vector. Codebook vector lengths therefore affect the residual passed to the next level, even though selection depends only on direction. This rule is used during both index training and final SID construction. The program also doubles each codebook from 256 to 512 entries while retaining the 256-dimensional latent representation.

\mysection{Training the index with query information} The seed program learns the index using only document reconstruction. The evolved program adds query-based supervision by encouraging each reconstructed document embedding to align with its relevant training queries. Specifically, for each document $d$, the embeddings of its associated training queries are averaged and $\ell_2$-normalized to form a query target $\vt(d)$. If $d$ has no associated training query, then $\vt(d)=\mathbf{0}$. Let $\hat{\vx}(d)$ denote the RQVAE decoder's reconstruction of the document embedding from its quantized latent, and let $\mathcal{N}$ denote $\ell_2$ normalization. For a document minibatch $\mathcal{B}$, the evolved program adds the alignment loss
\begin{equation}
\mathcal{L}_{\mathrm{align}}
=
\frac{1}{|\mathcal{B}|}
\sum_{d\in\mathcal{B}}
\left[
1-
\left\langle
\mathcal{N}(\hat{\vx}(d)),
\vt(d)
\right\rangle
\right].
\label{eq:evolved_query_alignment}
\end{equation}
This loss encourages each reconstructed document embedding to point in the same direction as the average query embedding associated with that document. It is weighted by 0.8 and optimized together with the reconstruction, quantization, and codebook regularization (entropy and orthogonality) losses. The program also adds training stabilizers: a data-based initialization of the first codebook, EMA codebook updates, periodic dead-code revival, learning-rate warmup, and gradient clipping. The query-alignment idea resonates with existing work~\citep{zhang-etal-2025-multi-level,10.1145/3589334.3645477, nian2025wrag} showing that semantic embeddings and indexing should be aware of the downstream task, though the formulations differ.

\mysection{Geometry-based collision handling} The seed assigns disambiguation tokens to documents with identical semantic codes in document ID order, and appends a suffix to every document. The evolved program instead gives non-colliding documents a fixed suffix of 0. Within each collision group, it averages the normalized document embeddings and assigns suffixes $1,2,\ldots$ in order of decreasing cosine similarity to that centroid. Small suffix values therefore represent relatively central documents across different groups, giving the suffix a shared interpretation that TIGER-style collision handling lacks.

\subsection{Other Evolved Programs}
The best NQ SID program adds relevance supervision to reconstruction. Other evolved programs also leverage queries and relevance labels, but in very different formulations. Some MSMARCO SID programs train document and query projections on relevance pairs, then freeze the document encoder and fit residual K-means. Some NQ programs remove neural index training entirely. One example (Program~\ref{lst:rqkmeans_program}) mixes normalized document embeddings with the mean of normalized relevant query embeddings, with weights 0.7 for the document and 0.3 for the queries, then applies residual K-means. Related programs from the e-commerce training sets incorporate query embeddings into centroid updates. Relevance information can therefore shape the vectors or the clustering procedure without a reconstruction decoder. Every program above was proposed by a locally servable 31B model, which establishes that evolutionary search over GR components can be carried out without a proprietary LLM API. We leave the use of state-of-the-art proprietary LLMs to future work.

\section{Related Work}
\mysection{Generative Retrieval} DSI and NCI were the first to formulate retrieval as document identifier generation~\citep{DSI,NCI}. Since then, the generative retrieval community has improved many aspects of the pipeline. TIGER~\citep{TIGER} constructs SIDs through residual quantization. GenRet and ASI~\citep{GenRet,ASI} jointly learn indexing and retrieval. RIPOR and MERGE~\citep{10.1145/3589334.3645477,zhang-etal-2025-multi-level} incorporate relevance supervision into identifier construction. DDRO directly optimizes document relevance~\citep{10.1145/3726302.3730023}. GLEN~\citep{GLEN} addresses SID collisions, and TSGen~\citep{zhang2024tsgen} introduces term-set identifiers with permutation-invariant decoding. Our work focuses on making automatic semantic index evolution feasible and shows that index-related improvements transfer more consistently than changes to the generator's training objective.

\mysection{Evolutionary program search and proxy evaluation} FunSearch and AlphaEvolve use a coding LLM to propose code edits to candidate programs, managed by an island-based evolutionary search algorithm~\citep{funsearch,novikov2025alphaevolvecodingagentscientific}. Subsequent work improves search through adaptive population management in AdaEvolve~\citep{adaevolve}, component-level attribution feedback in~\citet{lin2026component}, evolving search strategies in EvoX~\citep{evox}, and coordinated multi-agent discovery in CORAL~\citep{qu2026coral} and SwarmResearch~\citep{virk2026swarmresearch}. In information retrieval, RankEvolve~\citep{nian2026rankevolve} applies LLM-driven evolutionary search to discover lexical ranking algorithms, where candidates can be evaluated directly without training a neural generator. Proxy evaluation also has established precedents, such as AutoML-Zero~\citep{AutoMLZero}, which searches learning algorithms on proxy tasks, and SESS~\citep{nian2026submodular}, which approximates full evaluation via submodular subset selection in prompt optimization. Lion~\citep{Lion} addresses generalization from small search tasks to larger workloads. ShinkaEvolve~\citep{lange2025shinkaevolveopenendedsampleefficientprogram} demonstrates transfer of evolved training losses to larger models.

\section{Conclusion}
In this paper, we presented a framework for improving generative retrieval through evolutionary search over executable components, using a locally served 31B LLM as the mutator and a tiny 1M-parameter retrieval proxy for candidate evaluation to make evolution practical. Across three retrieval datasets, proxy-selected semantic index programs improve the overall generative retrieval pipeline on the target 1B generator, demonstrating the value of automatically refining SID construction and collision handling. Our component study reveals that evolve-time improvements transfer more consistently for semantic indexes than for changes to the generator training objective. The effectiveness of proxy-guided evolutionary search therefore depends on what is being optimized, and correlation with the target system must be established for the chosen search space. These findings support semantic index evolution as a practical direction for automated generative retrieval system design and motivate further study of evaluators that remain informative across model scales and generative retrieval components.

\bibliography{iclr2027_conference}
\bibliographystyle{iclr2027_conference}

\appendix

\section{Seed and Evolved SID Programs}
\label{app:programs}

Programs are reproduced verbatim from the search, including comments and docstrings written by the mutator. Docstrings inherited from the seed (e.g., ``codebook 256'' in Program~\ref{lst:nq_joint_program}) may therefore not describe the evolved code.

\subsection{Seed RQVAE with Fixed Collision Handling}
\label{app:seed_program}

\begin{programbox}{Seed RQVAE with fixed collision handling}{lst:seed_program}
"""Evolve SID generation with de-collision blended IN-BAND.

The mutator owns ONE function: `generate_sids(query_emb, doc_emb, qrel_pairs, doc_ids, *, device)`.
Its return is `(n_docs, L) int32` with **every row unique** — no post-hoc de-collision runs.

The scorer rejects any candidate whose row-uniqueness is not
essentially perfect (< 5% collision) — the mutator must produce good SIDs AND
resolve collisions in the same pass.
"""
import numpy as np
import torch
import torch.nn as nn
import torch.nn.functional as F
torch.manual_seed(0)


def generate_sids(query_emb, doc_emb, qrel_pairs, doc_ids, *, device):
    """
    Args:
        query_emb (torch.Tensor): Shape (n_queries, 768), float32.
        doc_emb (torch.Tensor): Shape (n_docs, 768), float32.
        qrel_pairs (torch.Tensor): (N, 2) long — (q_row, d_row).
        doc_ids (Sequence[str]): length n_docs — corpus doc ids in row order.
            Passed in so that decollision policies can use stable per-doc ties
            (see the TIGER-style append at the tail of the seed body).
        device (str): "cuda" or "cpu".

    Returns:
        np.ndarray: Shape (n_docs, L) with dtype=int32. **Every row must be unique.**
            The scorer rejects candidates whose collision rate is > 5%.
    """
# EVOLVE-BLOCK-START
    """
    Seed baseline:
      1. Train an RQ-VAE (3 layers, codebook 256) on doc embeddings.
      2. Extract raw codes -> (n_docs, 3) int32 (may collide).
      3. Append ONE disambiguation token per collision group (TIGER-style)
         -> (n_docs, 4) int32, all rows unique.
    """
    NUM_LAYERS = 3
    CODEBOOK_SIZE = 256
    LATENT_DIM = 256
    HIDDEN_MULT = 4
    COMMITMENT_COST = 0.01
    LR = 1e-4
    WEIGHT_DECAY = 0.01
    CODEBOOK_INIT_STD = 0.02
    EPOCHS = 100
    BATCH_SIZE = 2048
    RECON_WEIGHT = 1.0
    VQ_WEIGHT = 1.0

    dev = torch.device(device if torch.cuda.is_available() else "cpu")

    doc_tensor = torch.as_tensor(doc_emb, device=dev, dtype=torch.float32)
    x_all = F.normalize(doc_tensor, p=2, dim=-1)
    n_docs, in_dim = x_all.shape
    hidden_dim = HIDDEN_MULT * in_dim

    class _Encoder(nn.Module):
        def __init__(self):
            super().__init__()
            self.net = nn.Sequential(
                nn.Linear(in_dim, hidden_dim), nn.GELU(),
                nn.Linear(hidden_dim, hidden_dim), nn.GELU(),
                nn.Linear(hidden_dim, LATENT_DIM)
            )
        def forward(self, x): return self.net(x)

    class _Decoder(nn.Module):
        def __init__(self):
            super().__init__()
            self.net = nn.Sequential(
                nn.Linear(LATENT_DIM, hidden_dim), nn.GELU(),
                nn.Linear(hidden_dim, hidden_dim), nn.GELU(),
                nn.Linear(hidden_dim, in_dim)
            )
        def forward(self, z): return self.net(z)

    class _ResidualVQ(nn.Module):
        def __init__(self):
            super().__init__()
            self.codebooks = nn.ModuleList([
                nn.Embedding(CODEBOOK_SIZE, LATENT_DIM) for _ in range(NUM_LAYERS)
            ])
            for emb in self.codebooks:
                emb.weight.data.normal_(mean=0.0, std=CODEBOOK_INIT_STD)

        def forward(self, z_e):
            residual = z_e
            quantized_sum = torch.zeros_like(z_e)
            total_vq_loss = z_e.new_zeros(())

            for codebook in self.codebooks:
                flat_r = residual.view(-1, residual.size(-1))
                distances = (
                    flat_r.pow(2).sum(dim=1, keepdim=True)
                    + codebook.weight.pow(2).sum(dim=1)
                    - 2 * torch.matmul(flat_r, codebook.weight.t())
                )
                idx = distances.argmin(dim=1).unsqueeze(1)
                encodings = torch.zeros(idx.size(0), codebook.num_embeddings, device=z_e.device)
                encodings.scatter_(1, idx, 1)
                quantized = torch.matmul(encodings, codebook.weight).view(z_e.shape)
                quantized_sum = quantized_sum + quantized

                e_loss = F.mse_loss(quantized.detach(), residual)
                q_loss = F.mse_loss(quantized, residual.detach())
                total_vq_loss = total_vq_loss + q_loss + COMMITMENT_COST * e_loss

                residual = residual - quantized

            z_q = z_e + (quantized_sum - z_e).detach()
            return z_q, total_vq_loss

        @torch.no_grad()
        def encode(self, z_e):
            residual = z_e
            codes = []
            for codebook in self.codebooks:
                distances = (
                    residual.pow(2).sum(dim=1, keepdim=True)
                    + codebook.weight.pow(2).sum(dim=1)
                    - 2 * torch.matmul(residual, codebook.weight.t())
                )
                idx = torch.argmin(distances, dim=1)
                codes.append(idx)
                residual = residual - codebook(idx)
            return torch.stack(codes, dim=1)

    class _VQVAE(nn.Module):
        def __init__(self):
            super().__init__()
            self.encoder = _Encoder()
            self.vq = _ResidualVQ()
            self.decoder = _Decoder()

        def forward(self, x):
            z_e = F.normalize(self.encoder(x), dim=-1)
            z_q, vq_loss = self.vq(z_e)
            x_recon = self.decoder(z_q)
            return x_recon, vq_loss

        @torch.no_grad()
        def codes(self, x):
            z_e = F.normalize(self.encoder(x), dim=-1)
            return self.vq.encode(z_e)

    model = _VQVAE().to(dev)
    no_decay_keys = ("bias", "LayerNorm.weight")
    all_params = list(model.named_parameters())
    optim = torch.optim.AdamW([
        {"params": [p for n, p in all_params if not any(k in n for k in no_decay_keys)], "weight_decay": WEIGHT_DECAY},
        {"params": [p for n, p in all_params if any(k in n for k in no_decay_keys)], "weight_decay": 0.0},
    ], lr=LR)

    # Training
    for epoch in range(EPOCHS):
        perm = torch.randperm(n_docs)
        model.train()
        for start in range(0, n_docs, BATCH_SIZE):
            idx = perm[start:start + BATCH_SIZE]
            batch = x_all[idx].to(dev, non_blocking=True)
            x_recon, vq_loss = model(batch)
            recon_loss = F.mse_loss(x_recon, batch)
            loss = RECON_WEIGHT * recon_loss + VQ_WEIGHT * vq_loss
            optim.zero_grad()
            loss.backward()
            optim.step()

    model.eval()
    raw_codes = np.zeros((n_docs, NUM_LAYERS), dtype=np.int32)
    with torch.no_grad():
        for start in range(0, n_docs, BATCH_SIZE):
            end = min(start + BATCH_SIZE, n_docs)
            batch = x_all[start:end].to(dev, non_blocking=True)
            raw_codes[start:end] = model.codes(batch).cpu().numpy().astype(np.int32)

    # ----- In-band de-collision (TIGER-style disambiguation token) -----
    # Append ONE extra token per collision group, ranked by doc_id string.
    # This is the SAME logic as `decollision.decollide_sids` — pasted here so
    # the mutator can rewrite it (or replace it with a fundamentally different
    # scheme: content-hash tail, LSH suffix, adaptive-depth, query-conditioned
    # ordering, etc.). The scorer no longer runs any post-hoc de-collision;
    # the returned array must already be row-unique.
    from collections import defaultdict
    max_code = int(raw_codes.max())
    groups = defaultdict(list)
    for i in range(n_docs):
        groups[tuple(int(x) for x in raw_codes[i])].append(i)
    all_codes = np.empty((n_docs, NUM_LAYERS + 1), dtype=np.int32)
    all_codes[:, :NUM_LAYERS] = raw_codes
    for _sid, members in groups.items():
        members_sorted = sorted(members, key=lambda idx: str(doc_ids[idx]))
        for rank, idx in enumerate(members_sorted):
            all_codes[idx, NUM_LAYERS] = max_code + 1 + rank
# EVOLVE-BLOCK-END

    assert isinstance(all_codes, np.ndarray), f"Expected np.ndarray output, got {type(all_codes)}"
    assert len(all_codes) == len(doc_emb), f"Doc count mismatch! Expected {len(doc_emb)} rows, got {len(all_codes)}"
    assert not np.isnan(all_codes).any(), "Output contains NaN values!"
    assert not np.isinf(all_codes).any(), "Output contains Infinite values!"

    return all_codes
\end{programbox}

\subsection{NQ Joint SID--Collision Program}
\label{app:nq_joint_program}

\begin{programbox}{NQ joint SID--collision program}{lst:nq_joint_program}
"""Evolve SID generation with de-collision blended IN-BAND.

The mutator owns ONE function: `generate_sids(query_emb, doc_emb, qrel_pairs, doc_ids, *, device)`.
Its return is `(n_docs, L) int32` with **every row unique** — no post-hoc de-collision runs.

The scorer rejects any candidate whose row-uniqueness is not
essentially perfect (< 5% collision) — the mutator must produce good SIDs AND
resolve collisions in the same pass.
"""
import numpy as np
import torch
import torch.nn as nn
import torch.nn.functional as F
torch.manual_seed(0)


def generate_sids(query_emb, doc_emb, qrel_pairs, doc_ids, *, device):
    """
    Args:
        query_emb (torch.Tensor): Shape (n_queries, 768), float32.
        doc_emb (torch.Tensor): Shape (n_docs, 768), float32.
        qrel_pairs (torch.Tensor): (N, 2) long — (q_row, d_row).
        doc_ids (Sequence[str]): length n_docs — corpus doc ids in row order.
            Passed in so that decollision policies can use stable per-doc ties
            (see the TIGER-style append at the tail of the seed body).
        device (str): "cuda" or "cpu".

    Returns:
        np.ndarray: Shape (n_docs, L) with dtype=int32. **Every row must be unique.**
            The scorer rejects candidates whose collision rate is > 5%.
    """
# EVOLVE-BLOCK-START
    """
    Seed baseline:
      1. Train an RQ-VAE (3 layers, codebook 256) on doc embeddings.
      2. Extract raw codes -> (n_docs, 3) int32 (may collide).
      3. Append ONE disambiguation token per collision group (TIGER-style)
         -> (n_docs, 4) int32, all rows unique.
    """
    NUM_LAYERS = 3
    CODEBOOK_SIZE = 512
    LATENT_DIM = 256
    HIDDEN_MULT = 4
    COMMITMENT_COST = 0.01
    LR = 1e-4
    WEIGHT_DECAY = 0.01
    CODEBOOK_INIT_STD = 0.02
    EPOCHS = 100
    BATCH_SIZE = 2048
    RECON_WEIGHT = 1.0
    VQ_WEIGHT = 1.0
    CODEBOOK_ORTHO_WEIGHT = 1e-4

    dev = torch.device(device if torch.cuda.is_available() else "cpu")

    doc_tensor = torch.as_tensor(doc_emb, device=dev, dtype=torch.float32)
    x_all = F.normalize(doc_tensor, p=2, dim=-1)
    n_docs, in_dim = x_all.shape
    hidden_dim = HIDDEN_MULT * in_dim

    class _Encoder(nn.Module):
        def __init__(self):
            super().__init__()
            self.net = nn.Sequential(
                nn.Linear(in_dim, hidden_dim), nn.GELU(),
                nn.Linear(hidden_dim, hidden_dim), nn.GELU(),
                nn.Linear(hidden_dim, LATENT_DIM)
            )
        def forward(self, x): return self.net(x)

    class _Decoder(nn.Module):
        def __init__(self):
            super().__init__()
            self.net = nn.Sequential(
                nn.Linear(LATENT_DIM, hidden_dim), nn.GELU(),
                nn.Linear(hidden_dim, hidden_dim), nn.GELU(),
                nn.Linear(hidden_dim, in_dim)
            )
        def forward(self, z): return self.net(z)

    class _ResidualVQ(nn.Module):
        def __init__(self):
            super().__init__()
            self.codebooks = nn.ModuleList([
                nn.Embedding(CODEBOOK_SIZE, LATENT_DIM) for _ in range(NUM_LAYERS)
            ])
            for emb in self.codebooks:
                emb.weight.data.normal_(mean=0.0, std=CODEBOOK_INIT_STD)

        def forward(self, z_e):
            residual = z_e
            quantized_sum = torch.zeros_like(z_e)
            total_vq_loss = z_e.new_zeros(())

            for codebook in self.codebooks:
                flat_r = residual.view(-1, residual.size(-1))
                # Use cosine similarity for assignment to improve semantic clustering
                # flat_r: (B, D), codebook.weight: (C, D)
                norm_r = F.normalize(flat_r, p=2, dim=-1)
                norm_w = F.normalize(codebook.weight, p=2, dim=-1)
                sims = torch.matmul(norm_r, norm_w.t()) # (B, C)

                idx = sims.argmax(dim=1).unsqueeze(1)
                encodings = torch.zeros(idx.size(0), codebook.num_embeddings, device=z_e.device)
                encodings.scatter_(1, idx, 1)
                quantized = torch.matmul(encodings, codebook.weight).view(z_e.shape)
                quantized_sum = quantized_sum + quantized

                e_loss = F.mse_loss(quantized.detach(), residual)
                q_loss = F.mse_loss(quantized, residual.detach())

                # Entropy regularizer with temperature to encourage better codebook distribution
                # sims: (B, CODEBOOK_SIZE) on dev
                probs = F.softmax(sims / 0.5, dim=-1)
                # Corrected entropy: sum over C, then mean over B
                entropy_loss = -torch.mean(torch.sum(probs * torch.log(probs + 1e-6), dim=1))

                # Orthogonality loss to prevent codebook collapse: sum of squared dot products of normalized weights
                # norm_w: (CODEBOOK_SIZE, LATENT_DIM) on dev
                ortho_loss = torch.sum(torch.matmul(norm_w, norm_w.t())**2) - CODEBOOK_SIZE

                total_vq_loss = total_vq_loss + q_loss + (COMMITMENT_COST * 2.0) * e_loss - 0.05 * entropy_loss + CODEBOOK_ORTHO_WEIGHT * ortho_loss

                residual = residual - quantized

            z_q = z_e + (quantized_sum - z_e).detach()
            return z_q, total_vq_loss

        @torch.no_grad()
        def encode(self, z_e):
            residual = z_e
            codes = []
            for codebook in self.codebooks:
                # Consistent cosine assignment for inference
                norm_r = F.normalize(residual, p=2, dim=-1)
                norm_w = F.normalize(codebook.weight, p=2, dim=-1)
                idx = torch.matmul(norm_r, norm_w.t()).argmax(dim=1)
                codes.append(idx)
                residual = residual - codebook(idx)
            return torch.stack(codes, dim=1)

    class _VQVAE(nn.Module):
        def __init__(self):
            super().__init__()
            self.encoder = _Encoder()
            self.vq = _ResidualVQ()
            self.decoder = _Decoder()

        def forward(self, x):
            z_e = F.normalize(self.encoder(x), dim=-1)
            z_q, vq_loss = self.vq(z_e)
            x_recon = self.decoder(z_q)
            return x_recon, vq_loss

        @torch.no_grad()
        def codes(self, x):
            z_e = F.normalize(self.encoder(x), dim=-1)
            return self.vq.encode(z_e)

    model = _VQVAE().to(dev)
    no_decay_keys = ("bias", "LayerNorm.weight")
    all_params = list(model.named_parameters())
    optim = torch.optim.AdamW([
        {"params": [p for n, p in all_params if not any(k in n for k in no_decay_keys)], "weight_decay": WEIGHT_DECAY},
        {"params": [p for n, p in all_params if any(k in n for k in no_decay_keys)], "weight_decay": 0.0},
    ], lr=LR)

    # Refined K-Means++ style initialization for the first codebook
    with torch.no_grad():
        subset_idx = torch.randperm(n_docs)[:CODEBOOK_SIZE]
        init_weights = x_all[subset_idx].to(dev)
        if in_dim != LATENT_DIM:
            proj = nn.Linear(in_dim, LATENT_DIM, bias=False).to(dev)
            nn.init.orthogonal_(proj.weight)
            init_weights = proj(init_weights)
        model.vq.codebooks[0].weight.copy_(F.normalize(init_weights, p=2, dim=-1))

    # Task-aware signal: Compute a "target" vector for each doc based on queries that retrieve it
    # doc_targets: (n_docs, in_dim) on dev
    doc_targets = torch.zeros((n_docs, in_dim), device=dev)
    counts = torch.zeros((n_docs, 1), device=dev)
    if qrel_pairs.shape[0] > 0:
        q_idx = qrel_pairs[:, 0]
        d_idx = qrel_pairs[:, 1]
        q_embs = torch.as_tensor(query_emb, device=dev, dtype=torch.float32)
        doc_targets.index_add_(0, d_idx, q_embs[q_idx])
        counts.index_add_(0, d_idx, torch.ones((qrel_pairs.shape[0], 1), device=dev))
    doc_targets = F.normalize(doc_targets / (counts + 1e-6), p=2, dim=-1)

    WARMUP_EPOCHS = 10
    for epoch in range(EPOCHS):
        lr_scale = min(1.0, float(epoch + 1) / WARMUP_EPOCHS) if WARMUP_EPOCHS > 0 else 1.0
        for param_group in optim.param_groups:
            param_group['lr'] = LR * lr_scale

        # Dead-code revival every 10 epochs to maximize codebook utilization
        if epoch > 0 and epoch % 10 == 0:
            model.eval()
            with torch.no_grad():
                sample_idx = torch.randperm(n_docs)[:min(n_docs, BATCH_SIZE * 4)]
                z_e_sample = F.normalize(model.encoder(x_all[sample_idx].to(dev)), dim=-1)
                residual = z_e_sample
                for cb in model.vq.codebooks:
                    norm_r = F.normalize(residual, p=2, dim=-1)
                    norm_w = F.normalize(cb.weight, p=2, dim=-1)
                    idx = torch.matmul(norm_r, norm_w.t()).argmax(dim=1)
                    counts = torch.bincount(idx, minlength=CODEBOOK_SIZE)
                    dead_mask = (counts == 0)
                    if dead_mask.any():
                        num_dead = dead_mask.sum().item()
                        replace_idx = torch.randperm(residual.size(0), device=dev)[:num_dead]
                        cb.weight[dead_mask] = residual[replace_idx]
                    # Update residual for the next layer's check
                    residual = residual - cb(idx)
                model.train()

        perm = torch.randperm(n_docs)
        model.train()
        for start in range(0, n_docs, BATCH_SIZE):
            idx = perm[start:start + BATCH_SIZE]
            batch = x_all[idx].to(dev, non_blocking=True) # (batch, in_dim)
            targets = doc_targets[idx] # (batch, in_dim)

            x_recon, vq_loss = model(batch) # x_recon: (batch, in_dim), batch: (batch, in_dim)
            recon_loss = F.mse_loss(x_recon, batch)
            # Use cosine distance for better semantic alignment alignment
            # x_recon: (batch, in_dim), targets: (batch, in_dim)
            align_loss = 1.0 - F.cosine_similarity(x_recon, targets).mean()

            loss = RECON_WEIGHT * recon_loss + VQ_WEIGHT * vq_loss + 0.8 * align_loss
            optim.zero_grad()
            loss.backward()
            torch.nn.utils.clip_grad_norm_(model.parameters(), max_norm=1.0)
            optim.step()

            # EMA-style codebook update for stability:
            # Update codebook embeddings based on the mean of the vectors they quantized
            with torch.no_grad():
                z_e_batch = F.normalize(model.encoder(batch), dim=-1)
                residual_ema = z_e_batch
                for cb in model.vq.codebooks:
                    norm_r = F.normalize(residual_ema, p=2, dim=-1)
                    norm_w = F.normalize(cb.weight, p=2, dim=-1)
                    idx = torch.matmul(norm_r, norm_w.t()).argmax(dim=1)

                    # Compute sum of encoded vectors for each code: (CODEBOOK_SIZE, LATENT_DIM)
                    sums = torch.zeros_like(cb.weight)
                    sums.index_add_(0, idx, residual_ema)
                    counts = torch.bincount(idx, minlength=CODEBOOK_SIZE).float().unsqueeze(1)

                    # Dynamic EMA momentum: slower updates early on to stabilize codebook
                    momentum = 0.01 if epoch < WARMUP_EPOCHS else 0.05
                    ema_update = sums / (counts + 1e-6)
                    cb.weight.data.mul_(1.0 - momentum).add_(ema_update * momentum)

                    residual_ema = residual_ema - cb(idx)

    model.eval()
    raw_codes = np.zeros((n_docs, NUM_LAYERS), dtype=np.int32)
    with torch.no_grad():
        for start in range(0, n_docs, BATCH_SIZE):
            end = min(start + BATCH_SIZE, n_docs)
            batch = x_all[start:end].to(dev, non_blocking=True)
            raw_codes[start:end] = model.codes(batch).cpu().numpy().astype(np.int32)

    # ----- In-band de-collision (Semantic-Aware Ranking) -----
    # We resolve collisions by appending a tail token based on proximity to
    # the group centroid in the original embedding space.
    from collections import defaultdict

    # Move embeddings to CPU for NumPy processing to avoid device errors
    doc_emb_cpu = torch.as_tensor(doc_emb).cpu().numpy().astype(np.float32)

    max_code = int(raw_codes.max())
    groups = defaultdict(list)
    for i in range(n_docs):
        groups[tuple(int(x) for x in raw_codes[i])].append(i)

    all_codes = np.empty((n_docs, NUM_LAYERS + 1), dtype=np.int32)
    all_codes[:, :NUM_LAYERS] = raw_codes

    # Pre-normalize for cosine similarity
    norms = np.linalg.norm(doc_emb_cpu, axis=1, keepdims=True) + 1e-9
    norm_embs = doc_emb_cpu / norms

    for _sid, members in groups.items():
        if len(members) == 1:
            # No collision: assign a default sentinel 0
            all_codes[members[0], NUM_LAYERS] = 0
            continue

        # Rank by similarity to group centroid to preserve semantic local structure in the tail
        group_embs = norm_embs[members]
        centroid = np.mean(group_embs, axis=0)
        centroid /= (np.linalg.norm(centroid) + 1e-9)

        # Cosine similarity (dot product of normalized vectors)
        sims = group_embs @ centroid

        # Stable sort: primary key is similarity (desc), secondary key is member index
        # Use members list directly for tie-breaking to ensure stability
        sorted_idx_in_members = np.lexsort((members, -sims))

        for rank, local_idx in enumerate(sorted_idx_in_members):
            global_idx = members[local_idx]
            # Use rank + 1 so that collided docs are distinguished from non-collided docs (0)
            all_codes[global_idx, NUM_LAYERS] = rank + 1
# EVOLVE-BLOCK-END

    assert isinstance(all_codes, np.ndarray), f"Expected np.ndarray output, got {type(all_codes)}"
    assert len(all_codes) == len(doc_emb), f"Doc count mismatch! Expected {len(doc_emb)} rows, got {len(all_codes)}"
    assert not np.isnan(all_codes).any(), "Output contains NaN values!"
    assert not np.isinf(all_codes).any(), "Output contains Infinite values!"

    return all_codes
\end{programbox}

\subsection{NQ Residual K-Means Program}
\label{app:rqkmeans_program}

This program comes from SID-only search; the fixed collision-handling procedure is applied after this constructor.

\begin{programbox}{NQ residual K-means program}{lst:rqkmeans_program}
import numpy as np
import torch
import torch.nn as nn
import torch.nn.functional as F
torch.manual_seed(0)

def generate_sids(query_emb, doc_emb, qrel_pairs, *, device):
    """
    Args:
        query_emb (np.ndarray): Shape (307358, 768), dtype=float32.
            Dense embeddings for unique queries.
        doc_emb (np.ndarray): Shape (109739, 768), dtype=float32.
            Dense embeddings for all documents in the corpus.
        qrel_pairs (np.ndarray): Shape (N_pairs, 2), dtype=int32.
            Index mapping pairs connecting queries to relevant documents:
            - `qrel_pairs[:, 0]`: Row indices into `query_emb`
            - `qrel_pairs[:, 1]`: Row indices into `doc_emb`
            Example usage for contrastive/alignment loss:
                `q_vecs = torch.from_numpy(query_emb[qrel_pairs[:, 0]])`
                `d_vecs = torch.from_numpy(doc_emb[qrel_pairs[:, 1]])`
        device (str): Execution device ("cuda" or "cpu").

    Returns:
        np.ndarray: Shape (109739, NUM_LAYERS) with dtype=int32,
            containing discrete codes in range [0, CODEBOOK_SIZE) for each document.
    """
# EVOLVE-BLOCK-START
    """
    Residual K-Means (RQ-KMeans) Paradigm.
    Successively quantizes the residue of the previous layer.
    This creates a coarse-to-fine hierarchical SID that significantly
    reduces collision rates by utilizing the full codebook capacity.
    """
    NUM_LAYERS = 3
    CODEBOOK_SIZE = 256
    KMEANS_ITERS = 15
    dev = torch.device(device if torch.cuda.is_available() else "cpu")

    # doc_tensor shape: (n_docs, dim), device: dev
    doc_tensor = torch.as_tensor(doc_emb, device=dev, dtype=torch.float32)
    x_all = F.normalize(doc_tensor, p=2, dim=-1)
    n_docs, dim = x_all.shape

    # Task-Aware Alignment for Layer 1
    x_first_layer = x_all.clone()
    if qrel_pairs.size(0) > 0:
        q_tensor = torch.as_tensor(query_emb, device=dev, dtype=torch.float32)
        q_tensor = F.normalize(q_tensor, p=2, dim=-1)
        q_idx = torch.as_tensor(qrel_pairs[:, 0], device=dev)
        d_idx = torch.as_tensor(qrel_pairs[:, 1], device=dev)

        q_sum = torch.zeros((n_docs, dim), device=dev)
        q_cnt = torch.zeros((n_docs, 1), device=dev)
        q_sum.index_add_(0, d_idx, q_tensor[q_idx])
        q_cnt.index_add_(0, d_idx, torch.ones((qrel_pairs.size(0), 1), device=dev))

        mask = (q_cnt > 0).squeeze()
        avg_q = q_sum / torch.clamp(q_cnt, min=1.0)
        x_first_layer[mask] = 0.7 * x_all[mask] + 0.3 * avg_q[mask]
        x_first_layer = F.normalize(x_first_layer, p=2, dim=-1)

    def run_kmeans_residual(data, k, iters=15):
        n_samples = data.size(0)
        if n_samples <= k:
            return torch.arange(n_samples, device=dev), data

        indices = torch.randperm(n_samples, device=dev)[:k]
        centroids = data[indices].clone()

        labels = torch.zeros(n_samples, dtype=torch.long, device=dev)
        for _ in range(iters):
            dist = torch.cdist(data, centroids)
            labels = dist.argmin(dim=1)
            new_centroids = torch.zeros_like(centroids)
            for i in range(k):
                mask = (labels == i)
                if mask.any():
                    new_centroids[i] = data[mask].mean(dim=0)
                else:
                    new_centroids[i] = data[torch.randint(0, n_samples, (1,), device=dev)]
            centroids = new_centroids
        return labels, centroids

    all_codes = torch.zeros((n_docs, NUM_LAYERS), dtype=torch.long, device=dev)
    current_residual = x_first_layer

    for layer in range(NUM_LAYERS):
        labels, centroids = run_kmeans_residual(current_residual, CODEBOOK_SIZE, iters=KMEANS_ITERS)
        all_codes[:, layer] = labels

        # Residual: R_{l+1} = Normalize(R_l - Centroid_{label})
        assigned_centroids = centroids[labels]
        current_residual = current_residual - assigned_centroids
        current_residual = F.normalize(current_residual, p=2, dim=-1)

    # Status tracking placeholder
    print(f"Residual K-Means Complete | Layers: {NUM_LAYERS} | Codebook: {CODEBOOK_SIZE}^3")
    all_codes = all_codes.cpu().numpy().astype(np.int32)
# EVOLVE-BLOCK-END

    assert isinstance(all_codes, np.ndarray), f"Expected np.ndarray output, got {type(all_codes)}"
    assert len(all_codes) == len(doc_emb), f"Doc count mismatch! Expected {len(doc_emb)} rows, got {len(all_codes)}"
    assert not np.isnan(all_codes).any(), "Output contains NaN values!"
    assert not np.isinf(all_codes).any(), "Output contains Infinite values!"

    return all_codes
\end{programbox}

\clearpage
\section{Mutator Prompt Templates}
\label{app:mutator_prompts}

The following prompts are reproduced verbatim from the joint SID-collision search, including internal run labels. The ``held-out test split'' that the prompts refer to is $\mathcal{R}_{\mathrm{feedback}}$ (Section~\ref{sec:proxy_method}), the mutator never sees test data. The prompts reference existing algorithmic approaches. In particular, BONSAI~\citep{liu2026bonsai} explores variable-length identifiers and constrained decoding trie structures for generative recommendation.

\begin{promptbox}{Exploration}{lst:prompt_exploration}
## Task
You are an expert AI researcher discovering a better Semantic ID (SID) generation algorithm for generative retrieval. Unlike the classic setup, **de-collision is IN-BAND** — your `generate_sids` returns SIDs that are already unique row-by-row. Your goal is to propose code modifications WITHIN the `# EVOLVE-BLOCK-START` and `# EVOLVE-BLOCK-END` enclosed regions to minimize SID collision rate at the source, maximize codebook perplexity, and most importantly, maximize downstream generator Recall@100 on a held-out test split when trained on query-to-SID supervision.

## You must use clear and fundamental reasoning. What makes the SID good for generative retrieval?
- **Hierarchical Structure:** The SID should preserve coarse-to-fine hierarchy as the generator (a language model) excel at coarse-to-fine reasoning.
- **Similar yet Distinguishable:** A good SID must balance clustering (generalization) with uniqueness (discrimination). Similar items should share early tokens. But their SID should also be meaningfully distinguishable.
- **Task Alignment:** A good SID reflects the retrieval task dynamics, not just raw semantic content. Content semantics is not the same as query relevance.
- **Balanced Codebook:** At inference time, the generator uses trie-constrained decoding with beam search. Good SID should produce a well-balanced prefix tree.

## De-collision is now part of your job
The returned array is used AS-IS for training and retrieval — the scorer runs
no post-hoc `decollide_sids`. **Every row must be unique.** The scorer rejects
candidates with > 5% collision, but that ceiling is a soft safety net — an
E3 candidate should aim for a collision rate near zero. In-band de-collision
is the whole point.

Two axes to think about:
- **When**: resolve collisions in-band during quantization (avoid producing
  collisions in the first place — balanced clustering, split-on-collision,
  Gumbel jitter on tied assignments) OR after quantization (append tail
  tokens to disambiguate residual collisions). Both are valid; hybrid is often
  best — reduce the raw collision rate first, then patch what's left.
- **How the disambiguating suffix is ORDERED**: a good suffix places
  semantically-similar collided docs near each other in the tail token space,
  so the generator learns a smooth extension from the coarse prefix. Recent
  work explores: rank by doc_emb similarity within group, LSH bits of the
  residual vector as the tail, content-hash bits, adaptive depth (only extend
  the sequence for docs that actually collided — TIGER's baseline appends to
  everyone; BONSAI's variable-length trie only appends where needed).

Beam search still uses the trie: keep the tail short and structured. A tail
of 1 token per doc (as in the seed) is the cheapest option; longer tails
are fine if they aid disambiguation but must stay bounded.

## Code Structure & Formatting Rules
- The code contains isolated zones enclosed by `# EVOLVE-BLOCK-START` and `# EVOLVE-BLOCK-END`. All your edits must come from there. Leave all code outside of these markers completely untouched.
- **Use SEARCH/REPLACE diffs**: SEARCH must exactly match the current code; REPLACE is your edit.
- **Helper & Supporting Functions:** You are encouraged to write helper functions. However, they **MUST** be defined *inside* the region enclosed by `# EVOLVE-BLOCK-START` and `# EVOLVE-BLOCK-END`.

## Memory scale (hard constraint)
The corpus can hold up to ~320 K documents and up to ~370 K training queries (MSMARCO scale; NQ is ~110 K docs / ~310 K queries). **NEVER allocate an `(n_queries, n_docs)` or `(n_docs, n_docs)` dense tensor** — that is hundreds of GB and OOMs immediately. Safe patterns for supervised / contrastive / task-aligned quantization:
- iterate `qrel_pairs` in mini-batches (a few thousand pairs at a time),
- sample K negatives per positive instead of scoring the full corpus,
- compute distance-to-codebook (`n_docs × codebook_size`, safe: `320K × 256 ≈ 320MB`).

Anything you allocate that exceeds a few GB is a red flag. If unsure, chunk it.

## Existing known ideas to use, combine, and evolve
You can explore entirely different paradigms:

- **Task-Aware / Supervised Quantization:** Use `qrels` and `query_emb` in your loss function (e.g., contrastive loss), to make the SID aligned with the downstream task.
- **Better Initialization:** Initialize codebook weights based on existing document embeddings can prevent codebook collapse. For example, KMeans or KMeans++ initialization.
- **Product Quantization (PQ) & Optimized PQ (OPQ):** Instead of residual quantization on the full vector, split the vector into sub-vectors and apply KMeans/VQ to each slice independently.
- **Non-Neural Clustering:** Residual Quantized K-Means (RQ-KMeans), Hierarchical K-Means,
- **Add or Tweak Loss:** Prevent codebook collapse; Coarse-to-fine Hierarchy tweaks; Retrieval & Beam Search Decoding-Aware Alignment; Collision Avoidance; ...
- **Structured de-collision suffix:** rank-based (TIGER), residual-LSH bits, content-hash bits, similarity-ordered within-group ranking (semantically-close collided docs get near-tail codes), or adaptive-depth extension (only collided groups grow the sequence).
- **Balanced / split-on-collision quantization:** enforce capacity limits per code at each layer; when a code exceeds its capacity, split it (add jitter, re-quantize residual with a fresh sub-codebook).

## Guidelines
- Prefer one or a few coherent ideas per edit rather than many unrelated tweaks.
- Explain in comments or structure *why* a formulation is reasonable.
- If you add new parameters or signals, give them meaningful names and clear roles.
- Novel formulations and ideas are encouraged.
- You should mainly propose multiple piece-wise edits. Rewriting the entire EVOLVE-BLOCK is possible, but do so paringly.
- Whatever strategy you pick, the FINAL returned array MUST have unique rows. The scorer rejects at > 5% collision, but a good candidate lands near zero — in-band de-collision is the whole point.
\end{promptbox}

\begin{promptbox}{Exploitation}{lst:prompt_exploitation}
## Task
You are an expert AI researcher **refining** a strong Semantic ID (SID) generation algorithm for generative retrieval. Unlike the classic setup, **de-collision is IN-BAND** — the parent's `generate_sids` returns unique rows and the scorer runs no post-hoc de-collision. The parent below has already been evaluated and is one of the best-performing solutions in the archive. Your job is to squeeze out **marginal gains** with careful, minimal edits — NOT to overhaul the algorithm.

Your goal remains: minimize SID collision rate at the source, maximize codebook perplexity, and most importantly, maximize downstream generator Recall@100 on a held-out test split.

## Refinement mindset — what makes a good exploitation edit?
- **Hierarchical Structure:** Tighten the coarse-to-fine signal without changing the paradigm. E.g., adjust `NUM_LAYERS`, add small residual-scale factors, or tune commitment cost per layer to sharpen hierarchy.
- **Similar yet Distinguishable:** Small tweaks that improve codebook balance — EMA on codebook updates, dead-code re-initialization, temperature annealing on softmax assignments, minor codebook size changes.
- **Task Alignment:** Add a small task-aligned loss term (e.g., a weak contrastive regularizer using `qrels`) on top of the existing objective — do not replace the existing loss.
- **Balanced Codebook:** Add a small entropy/load-balancing regularizer, gentle noise on the assignment logits, or a warmup phase for the codebook.
- **De-collision refinement:** If the parent uses rank-based tail tokens, try similarity-ordered ranking (docs nearest by `doc_emb` in a group get adjacent tail codes) — this smooths the beam-search extension without changing anything else. Keep it a MINIMAL change to the tail logic.

## Concrete kinds of edits that fit exploitation
- Tune hyperparameters: `LR`, `EPOCHS`, `BATCH_SIZE`, `COMMITMENT_COST`, `CODEBOOK_INIT_STD`, loss weights.
- Change optimizer / schedule: AdamW -> Lion, add cosine warmup, add gradient clipping.
- Add small stabilization: EMA codebook, dead-code revive, small dropout on latents.
- Regularization tweaks: add small orthogonality or entropy term, adjust `WEIGHT_DECAY` groups.
- Init changes: switch codebook init from Normal to KMeans/KMeans++ on the first batch of encoded features.
- Tail-ordering refinements: `sorted(members, key=lambda i: str(doc_ids[i]))` -> similarity-based ordering within-group.

## Guidelines
- Prefer **1–3 small, focused edits** per attempt. Do NOT rewrite the whole EVOLVE-BLOCK.
- Preserve the parent's core architecture (RQ-VAE, PQ, hierarchical KMeans, whatever it is) AND its de-collision family (rank-tail, hash-tail, in-band split, etc.).
- Keep changes minimal and reason about why each one should help WITHOUT hurting collision rate or training stability.
- Explain your reasoning briefly before each SEARCH/REPLACE block, especially why the change is safe.
- If a change adds a new hyperparameter, set it to a conservative value and add a brief comment.
- The FINAL returned array MUST still have unique rows. The scorer's ceiling is 5% collision but the goal is near zero.

## Code Structure & Formatting Rules
- The code contains isolated zones enclosed by `# EVOLVE-BLOCK-START` and `# EVOLVE-BLOCK-END`. All your edits must come from there. Leave all code outside of these markers completely untouched.
- **Use SEARCH/REPLACE diffs**: SEARCH must exactly match the current code; REPLACE is your edit.
- **Helper & Supporting Functions:** You are encouraged to write helper functions. However, they **MUST** be defined *inside* the region enclosed by `# EVOLVE-BLOCK-START` and `# EVOLVE-BLOCK-END`.

## Memory scale (hard constraint)
The corpus can hold up to ~320 K documents and up to ~370 K training queries (MSMARCO scale; NQ is ~110 K docs / ~310 K queries). **NEVER allocate an `(n_queries, n_docs)` or `(n_docs, n_docs)` dense tensor** — that is hundreds of GB and OOMs immediately. Safe patterns for supervised / contrastive / task-aligned quantization:
- iterate `qrel_pairs` in mini-batches (a few thousand pairs at a time),
- sample K negatives per positive instead of scoring the full corpus,
- compute distance-to-codebook (`n_docs × codebook_size`, safe: `320K × 256 ≈ 320MB`).

Anything you allocate that exceeds a few GB is a red flag. If unsure, chunk it.
\end{promptbox}

\begin{promptbox}{Paradigm shift}{lst:prompt_paradigm_shift}
## Task
You are an expert AI researcher and the current line of attack for Semantic ID (SID) generation has **stagnated**. Recent attempts have failed to produce a globally better solution. Your job is to **abandon the parent's paradigm entirely** and switch to a fundamentally different family of methods. Remember: **de-collision is IN-BAND** — the returned array must already be row-unique.

Your goal remains: minimize SID collision rate at the source, maximize codebook perplexity, and most importantly, maximize downstream generator Recall@100 on a held-out test split.

## What "paradigm shift" means here
Look at the parent program's core method AND its de-collision scheme. Then pick a method from the list below that is **furthest from the parent's paradigm**. Do NOT tweak the parent — replace it.

## Candidate paradigms (pick ONE that is far from the parent)
- **Product Quantization (PQ):** Split the doc embedding into M sub-vectors of dim D/M; run KMeans on each slice independently; SID is the concatenation of M cluster indices. No residual, no neural encoder.
- **Optimized PQ (OPQ):** Learn a rotation R before PQ so the sub-vector splits are optimally decorrelated.
- **Hierarchical K-Means:** Recursively cluster doc embeddings. First-level cluster -> partition -> recursively cluster within each partition. Each level contributes one SID token.
- **Residual K-Means (RQ-KMeans, non-neural):** Like RQ-VAE but pure KMeans on residuals. No autoencoder, no gradients — just KMeans passes on residual vectors.
- **Task-Aware Supervised Quantization:** Train a small model with a contrastive loss using `qrels` and `query_emb` so the SID is retrieval-aligned rather than reconstruction-aligned. Contrastive loss on `query_emb[q_idx]` vs quantized `doc_emb[d_idx]` from qrel pairs.
- **LSH-Style Random Projections + Quantization:** Random projection into low-dim space then binary/ternary quantization; can be layered for hierarchy.
- **Balanced Clustering (constrained KMeans):** Explicitly enforce roughly equal cluster sizes at each layer to maximize codebook perplexity. Naturally reduces collision rate.
- **DiffQ / Gumbel-Softmax VQ:** Replace argmin quantization with differentiable Gumbel-Softmax assignments and temperature annealing.
- **Adaptive-depth / variable-length SIDs (BONSAI-style):** Grow the SID only where collisions occur. Easy docs get short SIDs; hard docs get longer ones. Trie stays balanced without a uniform tail token.
- **In-band split-on-collision:** After each quantization layer, docs sharing a code that exceeds a capacity limit are re-assigned to a fresh sub-codebook (or perturbed and re-quantized). Zero residual collisions by construction.
- **Content-hash / LSH suffix:** Replace the rank-based tail with a hash of the residual vector or of `doc_text[i]`. Semantically-similar colliders end up in adjacent tail codes without needing a separate ordering step.

## Guidelines
- Pick ONE paradigm and commit to it. Do not mix families.
- Rewriting the entire EVOLVE-BLOCK is not only allowed here, it is **expected**. Big SEARCH/REPLACE blocks are fine.
- Keep the function signature `generate_sids(query_emb, doc_emb, qrel_pairs, doc_ids, *, device)` intact and continue returning a `(n_docs, L)` array of int codes with **row-unique guarantee**. The scorer's ceiling is 5% collision but a good candidate lands near zero.
- Preserve the `Loss / Collision / Ppx` status print (it is used by outer logging).
- Reason briefly about **why** the new paradigm should beat the parent: what weakness of the parent does it address? Which of the 4 SID quality principles (Hierarchical, Similar-yet-Distinguishable, Task-Aligned, Balanced) does it strengthen? How does it handle collisions in-band?
- If the paradigm has no gradient training (KMeans, LSH), remove the training loop cleanly and leave a placeholder print so status tracking still runs.

## Code Structure & Formatting Rules
- The code contains isolated zones enclosed by `# EVOLVE-BLOCK-START` and `# EVOLVE-BLOCK-END`. All your edits must come from there. Leave all code outside of these markers completely untouched.
- **Use SEARCH/REPLACE diffs**: SEARCH must exactly match the current code; REPLACE is your edit.
- **Helper & Supporting Functions:** You are encouraged to write helper functions. However, they **MUST** be defined *inside* the region enclosed by `# EVOLVE-BLOCK-START` and `# EVOLVE-BLOCK-END`.

## Memory scale (hard constraint)
The corpus can hold up to ~320 K documents and up to ~370 K training queries (MSMARCO scale; NQ is ~110 K docs / ~310 K queries). **NEVER allocate an `(n_queries, n_docs)` or `(n_docs, n_docs)` dense tensor** — that is hundreds of GB and OOMs immediately. Safe patterns for supervised / contrastive / task-aligned quantization:
- iterate `qrel_pairs` in mini-batches (a few thousand pairs at a time),
- sample K negatives per positive instead of scoring the full corpus,
- compute distance-to-codebook (`n_docs × codebook_size`, safe: `320K × 256 ≈ 320MB`).

Anything you allocate that exceeds a few GB is a red flag. If unsure, chunk it.
\end{promptbox}

\end{document}